%% file: diseasespectrum.tex
\documentclass[a4paper,11pt]{article}
\usepackage{amsfonts,latexsym,epsfig,psfrag,amsmath,amssymb,url,subfig,color}
\include{userdefs}

\graphicspath{{./Figs/}}
\title{Models of Disease Spectra}
\author{Iead Rezek$\dag$ and Christian Beckmann$^\ddag$\\
$\dag$ Department of Engineering Science, \\
University of Oxford, U.K.\\
$\ddag$ MIRA Institute for Biomedical Technology and Technical Medicine, \\
University of Twente, NL.
\thanks{Email: i.rezek@imperial.ac.uk}}

\date{\today}
\begin{document}
\maketitle
\begin{abstract}

Case vs control comparisons have been the classical approach to the study of neurological diseases. However, most patients will not fall cleanly into either group. Instead, clinicians will typically find patients that cannot be classified as having clearly progressed into the disease state. For those subjects, very little can be said about their brain function on the basis of analyses of group differences. To describe the intermediate brain function requires models that interpolate between the disease states. We have chosen Gaussian Processes (GP) regression to obtain a continuous spectrum of brain activation and to extract the unknown disease progression profile. Our models  incorporate spatial distribution of measures of activation, e.g. the correlation of an fMRI trace with an input stimulus, and so constitute ultra-high multi-variate GP regressors. We applied GPs to model fMRI image phenotypes across Alzheimer's Disease (AD) behavioural measures, e.g. MMSE, ACE etc. scores,  and obtained predictions at non-observed MMSE/ACE values. The overall model confirmed the known reduction in the spatial extent of activity in response to reading versus false-font stimulation. The predictive uncertainty indicated the worsening confidence intervals at behavioural scores distance from those used for GP training. Thus, the model indicated the type of patient (what behavioural score) that would need to included in the training data to improve models predictions. 
\end{abstract}

\section{Introduction}


Functional Magnetic resonance imaging  (fMRI) has emerged as a powerful tool allowing us to significantly advance our understanding of various neurological conditions delineate them with ever finer precision.  To date, mental state and symptoms are increasingly understood as a continuum, ranging from mild and unspecific to severe and specific thereby reflecting subtle alterations in cause and/or effect. Alzheimer's Disease (AD) is a particular case in point. Alzheimer's disease is characterised by a progressive decline in memory and other cognitive abilities~\cite{McKhannG:84}. Implicated as an early stage of AD is Mild Cognitive Impairment (MCI), a condition associated with isolated memory loss distinct normal age-related memory loss.  Furthermore, AD severity levels are also being differentiated~\cite{NestorP:04}. Thus, far from being an isolated condition AD now regarded as a spectrum of subtly changing neuro-functional features, with MCI and mild AD as the earliest components~\cite{YetkinFZ:06}.

This every finer delineation of disease states has an obvious barrier: the statistical analysis. The majority of fMRI designs into neurological disorders  analyse ``case-control'' data with multiple disease stages. This approach identifies various disease states on the basis of patient symptoms and psychological exams. It subsequently compares structural or functional image markers between the disease groups. In AD, for example, the Mini-Mental State Exam and AddenbrookeÕs Cognitive Examination are common tools used to group the population into various sub-groups, such as patients with severe Alzheimer's Disease,  patients with Mild Cognitively Impairment (MCI) and normals. Each groups' (f)MR image phenotypes are subsequently compared to identify the various processes that distinguishes them.  

Case-control studies with multiple disease stages are advantageous for two reasons. First, they allow researchers to distinguish between the stages without the need to to assume anything about  the stages'  interdependencies. Second, the statistical analysis is well-understood and  ANOVA and t-test software isreadily available to quantify the differences between the observed image phenotypes. 

However, the very discretisation of the disease stages means the analysis is unable to recover any trends (with respect to behavioural markers) and cannot reveal any information about the relationship between the disease stages. Without an explicit model of the disease process, causal effects are difficult to establish and no predictions can be made for new cases, particularly when the bases for determining the disease stages are continuous diagnostic neuropsychological markers. 

Consequently there are no means to verify the disease model or provide vital reliability estimates, e.g. confidence intervals. Also, the formation of cohorts to allow group comparisons implies a) an information loss and b) and inefficient use of the available data. Since data collection involves serious expenditure, the inefficient use of data may lead to substantially increased research costs. 

We believe that Bayesian interpolation between the disease states is the key to obtaining simple, transparent and efficient models that can provide information about an individual patient's outcome.  In this paper, we have chosen Gaussian Processes (GP) regression~\cite{RasmussenCE:06} to obtain a continuous spectrum of brain activation as function of neuropsychological markers. GP regression models properly account for uncertainty in the measurements and allow a flexible design of interpolation profiles. Since they model continuous and non-linear functions they provide an unrestricted and high resolution description of disease processes. Crucially, Gaussian process models are more efficient in their use of data as they 
dispense with the traditional and information-erosive data binning stage of standard comparison approaches. Instead, they directly use all available data.

\subsection{Related Work}

In this work we interpolate fMRI image phenotypes across disease states, which in turn are defined though clinical behaviour scores, such as the Intelligence Quotient (IQ), Minimal Mental State Exam (MMSE) or Clinical Dementia Rating. Our aim is to monitor how brain function is altered as clinical scores indicate a gradual deterioration of the condition. This is the reverse of other studies~\cite{MarquandA:09,DeMartinoF:10,HuaienL:04,NiY:08,MarquandA:09} which predict behavioural scores (e.g. pain intensity) conditioned on image phenotype. This approach can be used to predict and validate behavioural scores whilst in our approach we aim to understand the alterations in neural processing with deteriorating health.  In effect, the role of the independent and dependent variables in our approach is the reverse to that of ~\cite{MarquandA:09}. Gaussian processes (GPs)  have also been applied to decode or predicting brain states~\cite{NiY:08} and for the quantification of  information content of brain voxels~\cite{MarquandA:10}. In these studies, the approach has been, again, to predict or classify with fMR images providing the multi-dimensional input variable and not the reverse. 

Relevance Vector Machine regression, radial basis function networks and longitudinal models are techniques that are closely related to Gaussian processes and have been applied to fMRI~\cite{DeMartinoF:10,HuaienL:04}. However, longitudinal models are parametric and so their predictions are biased by the assumed model, while radial basis function networks and RVMs  might produce predictions inconsistent with prior belief~\cite{RasmussenCE:05}. By contrast, GPR predictions are model free and consistent. Thus, the unique predictive qualities of  GP models hold great potential to support innovative clinical research.

\subsection{Paper outline}

We begin the our exposition with a description of the Gaussian Process regression (section~\ref{sec:Methods}), starting from first principles and concluding with the whole brain Gaussian process model. We subsequently apply Gaussian Process regression to simulated data, to illustrate the principles of the model, and then to fMRI data to obtain a continuous model of Alzheimer's Disease. Strengths, weaknesses and extensions of our approach are considered in Sections~\ref{sec:Discussion} and ~\ref{sec:Summary and Possibilities for Improvement}.

\section{Methods}
\label{sec:Methods}
\subsection{Principles of Gaussian Processes Regression}
\label{sec:Principles of Gaussian Processes Regression}

%
%
%

Gaussian processes (GPs)\cite{RasmussenCE:06,MacKay:2003,SteinML:99} are a powerful and flexible framework for performing Bayesian inference over functions. To see how GPs can be used to perform inference over functions, we begin by examining the finite multi-variate Gaussian distribution. 

\subsubsection{The Multi-variate Gaussian Distribution}

A simple bi-variate Gaussian distribution specifies a joint probability distribution of two potentially co-varying variables $y_{1}$ and $y_{2}$. Using vector notation in which the two variables are jointly denoted by $\mathbf{y}=\left(\begin{array}{c} y_{1}\\y_{2}\end{array}\right)$, the bivariate Gaussian distribution is given as
\begin{equation}
\begin{split}
P(\mathbf{y}|\mu,\Sigma) &= \frac{1}{(2\pi)^{N/2}}|\Sigma|^{-\frac{1}{2}}
\exp\left\{-\frac{1}{2} \left(\mathbf{y}-\mu\right)\tp \Sigma^{-1}\left(\mathbf{y}-\mu\right)\right\} \label{eqn:mvgauss}\\
& \equiv {\mathcal N}(\mathbf{y} | \mu,\Sigma)
\end{split}
\end{equation}
This distribution is characterised/conditioned on the mean vector  $\mu$ and the covariance matrix $\Sigma$. 

The bi-variate case is illuminating because it demonstrates the key properties of the multi-variate Gaussian distribution that carry through to the Gaussian Process. An important one is its marginalization property. The distribution of one of the variables is individually Gaussian, i.e. they are \emph{marginally} Gaussian
\begin{equation}
P(y_{1}|\mu',\Sigma')=\int P(y_{1}, y_{2}|\mu,\Sigma) d y_{2} =  {\mathcal N}(y_{1} | \mu',\Sigma')
\end{equation}

Furthermore the distribution of $y_{2}$ remains Gaussian when we learn the value of $y_{1}$ --- a property referred to as being \emph{conditionally} Gaussian. If we make the partitions of the covariance matrix $\Sigma$ explicit, 
\begin{equation}
\Sigma = 
\begin{bmatrix}
 \Sigma_{11} &  \Sigma_{12} \\  \Sigma_{21}\tp &  \Sigma_{22}
 \end{bmatrix}
\end{equation}
then the conditional probability of $y_{2}$ given $y_{1}$  
\begin{equation}
\begin{split}
P(y_{2}|y_{1},\mu'',\Sigma'')=\frac{ P(y_{1}, y_{2}|\mu,\Sigma)}{P(y_{1}|\mu',\Sigma')} =
&\frac{1}{(2\pi)^{N/2}}|\Sigma''|^{-\frac{1}{2}}
\exp\left\{-\frac{1}{2} \left(\mathbf{y}-\mu'' \right)\tp \Sigma''^{-1}\left(\mathbf{y}-\mu''\right)\right\}
\label{eqn:condbinorm}
\end{split}
\end{equation}
is Gaussian with mean $\mu''$ and covariance $\Sigma''$ given by the equations
\begin{eqnarray}
\mu'' =  & \mu_{2} -  \Sigma_{12} \Sigma_{22}\inv (y_{1}-\mu_{1}) 
\label{eqn:mvnpredmean}\\
\Sigma'' =  &  \Sigma_{11} -   \Sigma_{12}\Sigma_{22}\inv \Sigma_{21}
\label{eqn:mvnpredvar}
\end{eqnarray}
Equation~\eqnref{eqn:condbinorm} is especially important for prediction. By observing $y_{1}$, the variance of $y_{2}$ is  potentially reduced and  the mean of $y_{2}$ is shifted towards that of $y_{1}$, depending on the degree of co-variance between $y_{1}$ and $y_{2}$. Given two variables $y_{1}$ and $y_{2}$ with unit variance and correlation $\rho$, for example, the parameters $\mu''$ and $\Sigma''$ of the conditional distribution become
\begin{equation}
\begin{split}
\mu'' =  & \mu_{2} -  \rho (y_{1}-\mu_{1}) \\
\Sigma'' =  &  1 -   \rho ^{2}.
\end{split}
\end{equation}
Thus, if $y_{1}$ and $y_{2}$ are correlated, observing  $y_{1}$ informs us about $y_{2}$ and vice versa.

\subsubsection{From Gaussian Distributions to Gaussian Processes}


We can extend the properties of multivariate Gaussian distributions to Gaussian processes with infinitely many dimensions. To prevent an explosion in the number of covariance parameters, the covariance between variables is defined by a function with only a handful of parameters.  For instance, a typical  covariance function, which must produce a positive semi-definite covariance matrices, expresses that variables with similar indices have strong correlations and those with dissimilar indices have weak correlations. In this case the closer an unknown variable is to an observed variable the lower its uncertainty. This concept remains valid for an infinite continuum of variables, with the expectation becoming a smooth function fitting the observations and the uncertainty increasing smoothly with distance to the closest observation. 

Equation~\eqref{eqn:mvgauss} provides the general matrix form for the probability density in a multi-variate Gaussian distribution. To extend this finite collection of variables to a function-valued variable, which represents an infinite number of variates, assume that we have observed a function $y(x)$ at a finite set of sampling points, $\mathbf{x}$, with the observed function values being $y(\mathbf{x})$. A Gaussian Process is then defined~\cite{RasmussenCE:06} as a collection of random variables, any finite number of which have a joint Gaussian distribution. This implies that if $Y$ is a function-valued random variable that has a GP distribution, any finite subset of $Y$ also has a Gaussian distribution and, therefore,
\begin{equation}
\begin{split}
p(Y = y(\mathbf{x}) | \theta) = N\left(y(\mathbf{x});m(\mathbf{x};\theta),k(\mathbf{x},\mathbf{x};\theta)\right)
\end{split}
\end{equation}
where $m(\mathbf{x};\theta)$ and $k(\mathbf{x},\mathbf{x};\theta)$ are the mean and covariance \emph{functions} of the GP that are parameterised by $\theta$, and $k(\mathbf{x},\mathbf{x};\theta)$ represents the matrix evaluation of the $k(x, x')$ for all possible pairs of elements of the vector $\mathbf{x}$. The vector $\mathbf{x}$ are commonly taken as the ``inputs'' of the GP and the values $\mathbf{y}$ as the outputs. 

In this case the full collection of random variables are all the values of $y(x)$, which is equivalent to a vector of infinite length, with one element for every potential input value. The finite collection are the sampled values, $y(\mathbf{x})$. As per the above definition we can model the entire function by assuming that any finite set of samples obeys a multi-variate Gaussian distribution. The mean, $m(x)$, is simply a function that describes the expected value of $y(x)$ before we make observations, and the mean for a finite set is simply $m(\mathbf{x})$.  The covariance kernel is defined over all potential pairs of inputs, and represents the covariance of function output values in terms of their inputs,
\begin{equation}
Cov\left(y(x_{i}),y(x_{j})\right)=\Sigma_{ij} =k(x_{i},x_{j};\theta)
\label{eqn:kernelfun}
\end{equation}
Like its finite dimensional counterpart, the covariance kernel must be a positive semi-definite function, meaning that any set of inputs must produce a covariance matrix, $\Sigma =k(\mathbf{x})$, with elements given by equation~\eqref{eqn:kernelfun}, must not have negative eigenvalues.  Colloquially, this means that if one function value is highly correlated with a second, which is in turn highly correlated with a third, there must also exist a significant correlation between the first and third values. It also means that uncertainty can not be negative. 

The mean and covariance functions define the types of outputs we expect to see from the GP distribution. The mean specifies what we expect the function output values to be, prior to making any observations. In many problems, a popular choice is  to set $m(x) = 0$. However, the mean may naturally be specified depending on the problem at hand.  The particular functional form of the covariance function is subject to restrictions to ensure the positive semi-definite requirement for the co-variance matrix but also encodes, among other things, the function's smoothness, amount of variability, and degrees of differentiability. The covariance can also incorporate more detailed knowledge of the function, such as potential periodicity or the existence of Ôchange-pointsÕ Ñ where the function values exhibit little or no relation across a boundary. 

As mentioned above, the covariance function expresses the dependence of pairs of output values as a function of their respective input values. In the most widely used examples this is simply a function of the separation between the input values, 
\begin{equation}
Cov\left(y(x_{i}),y(x_{j})\right)= k(|| x_{i},x_{j}||;\tau,\lambda)
\end{equation}
Of these, the squared exponential covariance with parameters $\tau$ and $\lambda$
\begin{equation}
k_{SE}(|| x_{i},x_{j}||;\tau,\lambda)= \lambda^{2} \exp\left(\frac{-1}{2\tau^{2}} (x_{i}- x_{j})\tp (x_{i}- x_{j}) \right)
\label{eqn:SE}
\end{equation}
is a standard and highly popular choice of covariance function.  The degree of smoothness on the output values, in the case of the  squared exponential covariance, is controlled by the input scale parameter $\tau$. Larger input scales increase the range of correlations as a function of the input value separation, leading to greater coupling between distant output values. This makes the resulting functions smoother and more slowly varying. Conversely, smaller input scales  lead to more rapidly varying functions where output values become uncorrelated with smaller differences in inputs. The parameter $\lambda$ controls the output scale. It simply acts as a rescaling factor for the entirety of the output function.


\subsubsection{Interpolation with Gaussian Processes}

Gaussian process kernels are frequently used to specify prior distributions over unknown functions which are observed at discrete points $\mathbf{x}$. The mathematically most convenient form for likelihood of the observation is the Gaussian distribution with zero mean and independent and identically distributed noise with variance $\sigma^{2}$, so that the the log-marginal distribution, $ \mathcal{L}\equiv\log\left(p(y\left(\mathbf{x}) |\theta\right)\right) $,  of the observed function values is given by 
\begin{equation}
 \mathcal{L} = -\frac{N}{2}\log(2\pi) - \frac{1}{2} | k(\mathbf{x},\mathbf{x}) | - \frac{1}{2} \left(\mathbf{y}-m(\mathbf{x})\right)\tp k(\mathbf{x},\mathbf{x})^{-1} \left(\mathbf{y}-m(\mathbf{x})\right)
\label{eqn:loggplike}
\end{equation}
where we have assumed the number of observations to be $N$ and where $ k(\mathbf{x},\mathbf{x};\theta) $ is now a composite of the squared exponential covariance function, function~\eqref{eqn:SE} and the observation noise
\begin{equation}
k(\mathbf{x},\mathbf{x};\theta) = k_{SE}(|| x_{i},x_{j}||;\tau,\lambda) + \sigma^{2}
\label{eqn:SEplusnoise}
\end{equation}

We are interested in making predictions about the value of the function $y(x_{*})$ at some new input $x_{*}$. In analogy to the bi-variate case, equations~\eqref{eqn:mvnpredmean}
and~\eqref{eqn:mvnpredvar}, the distribution of function values $y(x_{*})$ at $x_{*}$ is given as 
\[
p\left( y(x_{*}) | y(\mathbf{x})\right) = N \left(y(x_{*}); \mu_{*}, \Sigma_{*}\right) ,	
\]
with the updated mean and covariance matrices:
\begin{eqnarray}
\mu_{*} & = m(x_{*})+k(x_{*},\mathbf{x}) k(\mathbf{x},\mathbf{x};\theta)\inv (\mathbf{x}-m(\mathbf{x}))
\label{eqn:gppredmean} \\
\Sigma_{*} & = k(x_{*},x_{*}) - k(x_{*},\mathbf{x})k(\mathbf{x},\mathbf{x};\theta)\inv k(\mathbf{x},x_{*})
\label{eqn:gppredvar}
\end{eqnarray}


\subsection{Gaussian Process Regression for Disease Modelling}

Our approach is to use Gaussian process regression (GPR)~\cite{RasmussenCE:06} to obtain a description of the continuous spectrum of brain activation associated with behavioural markers~\cite{RezekI:10,NiY:08}. In their na\"{i}ve form, voxel-wise basis Gaussian process regression can be applied to observed image phenotypes values as a function of behavioural markers. Such models, however, will not take account the spatial dependencies between neighbouring image voxels. While sparse spatio-temporal GPR could provide a solution to this problem, our approach imposes spatial constraints on the hyper-parameters of the Gaussian Process. In particular, given the log-marginal likelihood of the observations at voxel $v$, 
\begin{equation}
 \log\left(p(y_{v}\left(\mathbf{x}) |\theta\right)\right)  = -\frac{N}{2}\log(2\pi) - \frac{1}{2} \log | k(\mathbf{x},\mathbf{x};\theta) | - \frac{1}{2} \left(\mathbf{y}_{v}-m(\mathbf{x})\right)\tp k(\mathbf{x},\mathbf{x};\theta)^{-1} \left(\mathbf{y}_{v}-m(\mathbf{x})\right)
\label{eqn:voxloggplike}
\end{equation}
the probability of the log-hyper parameters, $\ell \equiv \log(\theta)$, is assumed to follow a multivariate conditional autoregressive model (CAR)~\cite{BesagJ:74}
\begin{equation}
p\left(\ell_{v}|\ell_{-v}; \mathbf{B}_{v,-v}, \mathbf{T}_{v}\right) = N\left(\textstyle\sum_{v'\in \mathcal{N}(v)}\mathbf{B}_{v,v'} \, \ell_{v'}\, , \mathbf{T}_{v}\right),
\label{eqn:mvcar}
\end{equation}
where $-v$ denotes the indices of the voxels neighbouring voxel $v$,  $\mathbf{B}_{v,v'}$ is a $p \times p$ matrix ($p$ is the dimensionality of $\theta$) and $\mathbf{T}_{v}$ is a $p \times p$ covariance matrix. We assume that $\mathbf{B}_{v,v'} T_{v'} = T_{v} \mathbf{B}_{v',v}\tp$, to ensure symmetry, and  that $\mathbf{B}_{v,v}=-I$, where $I$ is a $p \times p$ identity matrix, to ensure the positive definiteness of the joint distribution $p\left(\ell; \mathbf{B}, \mathbf{T}\right)$ (here $\mathbf{B}=\{\mathbf{B}_{v,v'}\};\mathbf{T}=\{\mathbf{T}_{v}\}, \forall v$). Further, we make the additional simplifying assumptions that the covariance matrices are diagonal and voxel independent, 
\begin{equation}
\mathbf{T}_{v}\equiv \mathbf{T}_{0} = \left(\begin{array}{ccc} t_{1} & 0 & \cdots\\
0 &\ddots& 0\\
\cdots & 0 & t_{p} \end{array}\right),
\end{equation}
thus resembling the spatial smoothing assumptions underpinning the usual fMRI preprocessing steps. Also, we assume 
\begin{equation}
\mathbf{B}_{v,v'}\equiv \mathbf{B}_{0}=\frac{1}{|\mathcal{N}(v)|}  \mathbf{R},
\end{equation}
where $|\mathcal{N}(v)|$ is the number of neighbours of $v$ and the matrix $\mathbf{R}$ consists of constant weights  
\begin{equation}
\mathbf{R}= \left(\begin{array}{ccc} \rho_{1} &  \cdots & \rho_{1}\\
 &\ddots& \\
\rho_{p} & \cdots  & \rho_{p} \end{array}\right).
\end{equation}

In summary, in the full model the probability of the observed image features is
given by
\begin{equation}
p(y,l|\mathbf{B}_{0}, \mathbf{T}_{0}) = \prod_{v}  p(y_{v}\left(\mathbf{x}) | \ell_{v}\right) p\left(\ell_{v}|\ell_{-v}; \mathbf{B}_{0}, \mathbf{T}_{0}\right),\label{eqn:wholebrainpost}
\end{equation}
where
\begin{equation}
\begin{split}
\log\left(p(y_{v}\left(\mathbf{x})| \ell_{v}\right)\right)  =  -\frac{N}{2}\log(2\pi) - & \frac{1}{2} \log | k(\mathbf{x},\mathbf{x};\ell_{v}) |  \\ & - \frac{1}{2} \left(\mathbf{y}_{v}-m(\mathbf{x})\right)\tp k(\mathbf{x},\mathbf{x};\ell_{v})^{-1} \left(\mathbf{y}_{v}-m(\mathbf{x})\right) \end{split}
\end{equation}
and
\begin{equation}
\begin{split}
 \log\left(p\left(\ell_{v}|\ell_{-v}; \mathbf{B}_{0}, \mathbf{T}_{0}\right)\right) =&  -\frac{p}{2}\log(2\pi)  - \frac{1}{2}\log  | \mathbf{T}_{0}| \\ &- \frac{1}{2} \left(\left(\ell_{v}-\textstyle \sum_{v'\in \mathcal{N}(v)}\mathbf{B}_{0} \; \ell_{v'}\right)\tp T_{0}\inv \left(\ell_{v}-\textstyle \sum_{v'\in \mathcal{N}(v)}\mathbf{B}_{0}\;  \ell_{v'}\right)  \right)
 \end{split}
\end{equation}
Thus, every voxel's GP hyper-parameters are coupled to those of its immediate neighbours\footnote{The neighbourhood system is throughout assumed to be a first order one, i.e. its neighbours are those voxels that lie one voxel to the left, right, top, bottom, up and below the current voxel $v$.}. As hyper-parameters distributions have support only on the positive axis, we assume that they are log-normally distributed and that each log-hyper-parameter expectation is a voxel independent, weighted and scaled average of its neighbours with constant variance $\mathbf{T}_{0}$ and weighting $\mathbf{B}_{0}$. 

\subsection{Parameter Estimation}

Theoretically, there are a number of possible ways of estimating the parameters of the model~\eqref{eqn:wholebrainpost}. One feature that guides the possible choices are the non-linearities in the model, within the covariance function, and the non-conjugacies, due to the log-normal distribution assumptions of the hyper-parameters. While Markov Chain Monte Carlo methods~\cite{NealR:97} are applicable, they were ruled out early on because of their computational cost. This cost is too high considering that for a normal resolution fMRI some $10^{5}$ coupled Gaussian processes regression models require calibration. An alternative inference method is Expectation Propagation~\cite{RasmussenCE:06}(EP). EP does offer a viable and approximate Bayesian estimation of the model. However, a downside of EP is the additional integration that would have to be performed when computing the posterior predictive distributions for the purpose of interpolation.  

In this work we have opted for maximum \emph{aposteriori} estimation of the model parameters using an iterative procedure, known as iterated conditional modes (ICM). As the joint probability~\eqref{eqn:wholebrainpost} is difficult to maximise analytically, Besag~\cite{BesagJ:86}  proposed the iterated conditional modes algorithm. It maximises the posterior by sequentially maximising the local conditional probabilities
\begin{equation}
p\left(\ell_{v} | y_{v}(\mathbf{x}), \ell_{-v}\right)  \propto p\left(y_{v}(\mathbf{x}) | \ell_{v}\right) p\left(\ell_{v}|\ell_{-v}; \mathbf{B}_{0}, \mathbf{T}_{0}\right). 
\label{eqn:icm}
\end{equation}
For a given value of $\ell_{-v}^{k-1}$ at iteration $k-1$, the iteration seeks the value of $\ell_{v}$ which maximises~\eqref{eqn:icm} at the current iteration, $k$, 
\begin{equation}
\ell_{v}^{k} =\argmax{\ell_{v}}\log\left(p\left(\ell_{v} | y_{v}(\mathbf{x}), \ell_{-v}^{k-1}\right)\right).
\label{eqn:icmargmx}
\end{equation}
and continues until convergence. The convergence is guaranteed (for the serial updating), is rapid~\cite{BesagJ:86} but also depends very much on the initialisation.  These initialisation values are computed by calibrating the voxel-wise Gaussian processes, equation~\eqref{eqn:voxloggplike}, to the data using the conjugate gradient method that is part of the Gaussian process toolbox \texttt{GPML}~\cite{RasmussenCE:06}.  

Following initialisation, the procedure selects each voxel at random and updates the maximum \emph{aposteriori} parameter values, $\ell_{v}$, conditioned on the maximum values of its neighbours. These updates are calculated for all the voxels, $v=1,\cdots V$, in the volume. Several such sweeps, $k=1,\cdots, K$, over the entire volume are performed. However, we have empirically noticed that only a few sweeps ($K=5$) are required for estimation, which is in line with the experience of ICM reported in the literature.

\section{Results}
\label{sec:Results}

\subsection{Synthetic Illustrative Example}
\label{sec:Synthetic Illustrative Example}

Before describing the full model it instruct to  demonstrate Gaussian Process interpolation on a simulated population image example.  The simulated population consisted of $7$ subjects which had, respectively, the normalised behavioural scores $\{0,0.1, 0.3,0.7,0.8,0.9,1\}$. The extreme ends of the score range represents the clear states of a disease, i.e. the class of ``healthy'' individuals on one end and the class of patients on the other end of the scale.  
Each individual's underwent an imagined experimental protocol, consisting of $8$ repetitions of a stimulus that lasted for $48$ time-points. The generation process of the actual images resulting from this simulated protocol follows that described in~\cite{LangeN:99}.

\subsubsection{Image Generation}


The brain images used for simulation were based on a set of whole-brain images collected from a healthy volunteer was asked simply to relax and remain still while 73 whole-brain baseline scans were collected (each consisting of $30$ $2.19\times 2.19 \times 3.25-mm$ slices  collected using a Phillips $3.0T$ scanner with an echo-planar imaging (EPI) sequence  and TR=$2s$, TE =$30ms$).  To create the simulated time series corresponding to eight runs of 48 time-points each fMRI image sequence was resampled with repetition. To expedite the simulation, the effects of a simple on/off reference function operating in each run were constrained to a $12\times 24$-pixel patch around the central sulcus. 

Each pixel time series was linearly de-trended by removing a linear regression line, i.e., replacing the series with residuals from a least-squares fit of the intercept and slope parameters for each run, within each time series, separately. Each image voxel was subsequently standardised to unit-variance. The degree of activity at every voxel, $v$, is simulated by adding to the base-line series a periodic binary reference function $x$ (each period has $12$ time-steps off, $24$ on, and $12$ off) of  magnitude $\beta_{v}$. For reasons outlined in~\cite{LangeN:99}, this magnitude is specified some positive fraction, $m$, of  fMRI signal's standard deviation at location $v$,  $\sigma_{v}$,
\[
\beta_{v}= m \sigma_{v} 
\]  
The fraction $m$ of the standard deviation used in our experiments is set at unity.
\begin{figure}[htbp]
\begin{center}
\includegraphics[width=0.48\textwidth]{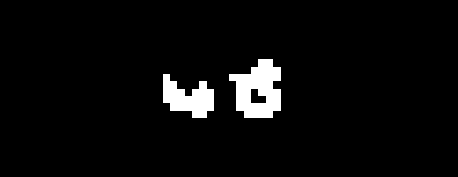}
\caption{Ground truth image for the active region in the simulation of a Gaussian process prediction of z-scores given a hypothetical behavioural measure (normalised between $0-1$).}
\label{fig:IonIoff}
\end{center}
\end{figure}

To create a dependence of the fMRI signal on the behavioural score,  two regions of activity are defined where were turned on or off depending on the behavioural score. These regions of ÔÔground truthÕÕ activity are shown in figure~\ref{fig:IonIoff} and referred to as region ${\mathcal A}$ and ${\mathcal B}$. The degree of added activity in each region was weighted  depending on on the normalised behavioural scores. More precisely, if voxel $v$ is in region ${\mathcal A}$, written as $\delta\left( v \in {\mathcal A}\right)$ where $\delta\left(\cdot\right)$ is the indicator function, and the subject has the behaviour score $\alpha$ then
\begin{equation}
\beta_{eff}=\alpha \delta\left( v \in {\mathcal A}\right) \beta_{v} + (1-\alpha) \delta(v \in {\mathcal B}) \beta_{v}
\end{equation}
is the effective magnitude of the signal at $v$  that is added to the base-line fMRI time series. Thus, the degree of activity is gradually shifted with one ground truth region (Region ${\mathcal A}$) active at higher normalised behavioural scores and the other region at lower scores.

\subsubsection{Gaussian Process Regression}

As mentioned above, the simulated image data was generated for $N=7$ behavioural scores, representing $7$ subjects with unique neuro-psychological assessment values. For each subject, a voxel-wise general linear model was fitted, resulting in parameter estimate $\hat{\beta}_{v}$ for voxel location $v$. This estimate was subsequently converted to a z-score and so the data is reduced to one z-score for each voxel.  A voxel-wise Gaussian process model, with squared-exponential plus noise covariance function,  was then adapted to the image features. 

\begin{figure}[htbp]
\begin{center}
\subfloat[The predictive mean]{\label{fig:imageinterp}\includegraphics[width=0.48\textwidth]{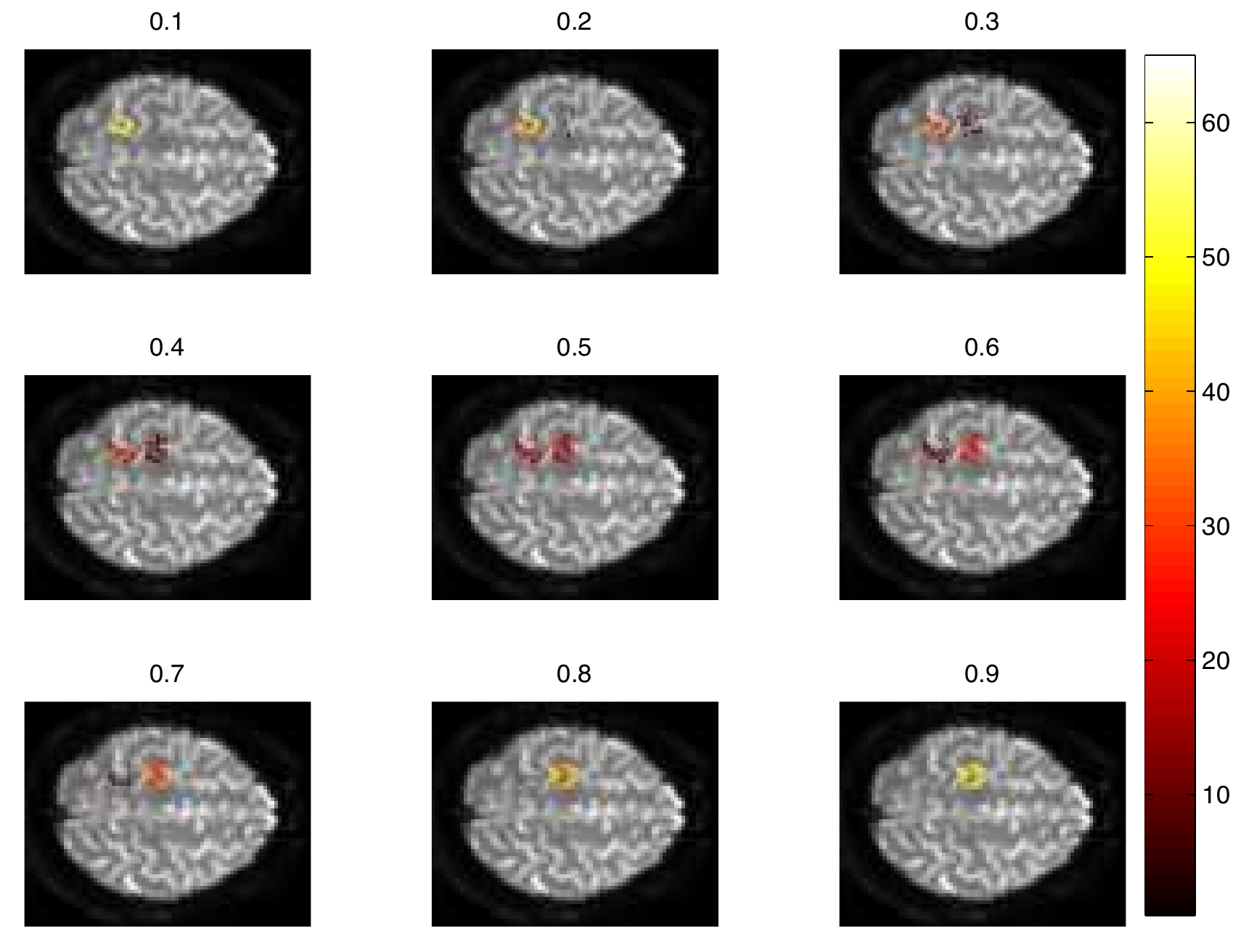}}\quad
\subfloat[The predictive variance]{\label{fig:imagevarinterp}\includegraphics[width=0.48\textwidth]{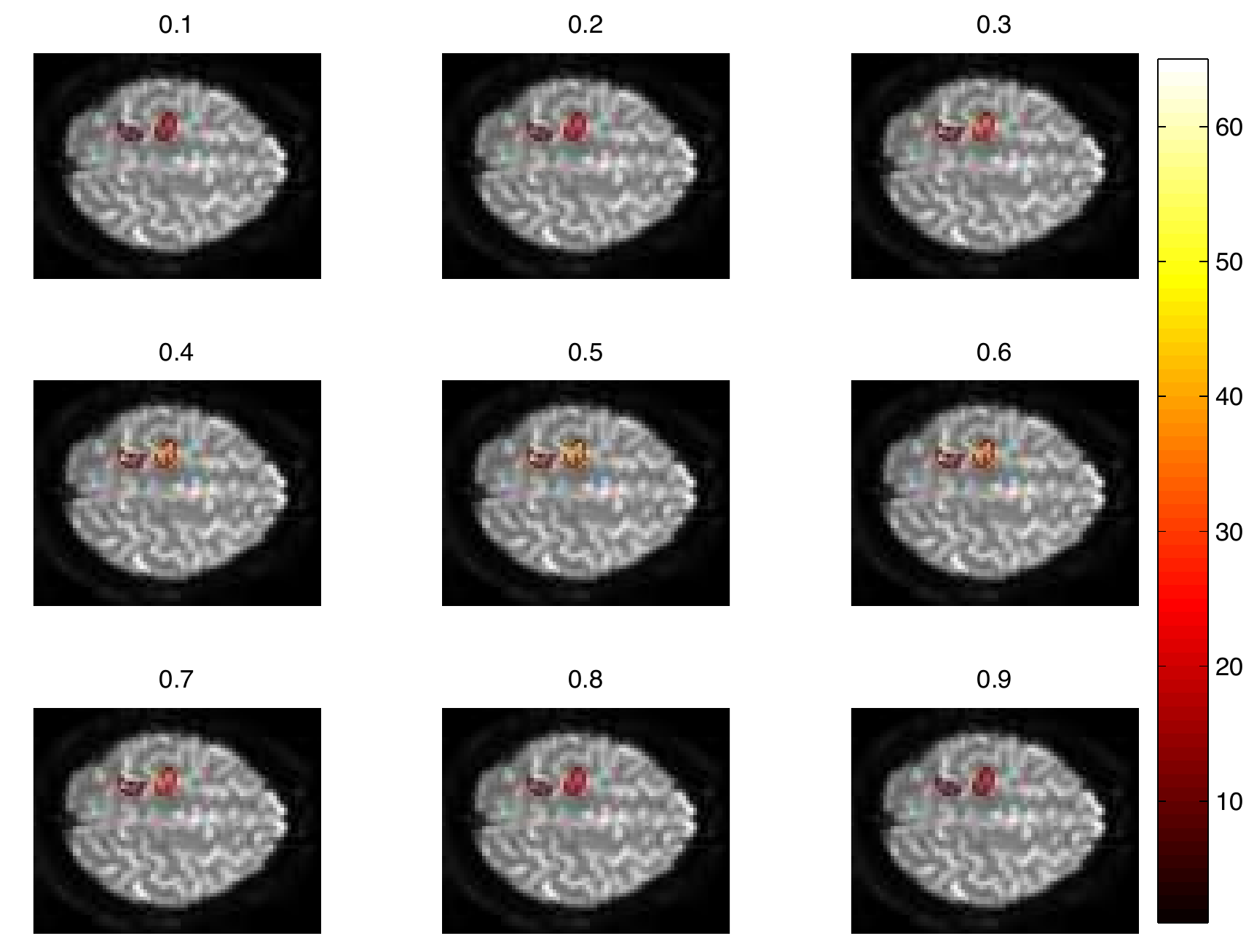}}\quad  
\caption{Simulation of a Gaussian process prediction of z-scores given a hypothetical behavioural measure (normalised between $0-1$).}
\label{default}
\end{center}
\end{figure}

The GPs were evaluated at the unobserved factor values ($0.4,0.5,0.6$) to produce posterior mean and variance images.  E.g. at values around $0.5$, see figure~\eqref{fig:imageinterp}, the model suggests two active regions. However, the variance image, see figure~\eqref{fig:imagevarinterp} also indicates a high degree of uncertainty. This variance may be used to guide future fMRI experiments. A confirmation of an actual measurement at factor value of $0.5$, say, would suggest a gradual change of cognitive function with behaviour, whilst a refutation suggests a sudden transition of function. 

\subsection{Alzheimer's Disease Data}
\label{sec:Alzheimer's Disease Data}
\subsubsection{Participants and fMRI procedures}
We analysed data  collected as part of an Alzheimer's Disease study. Fifteen participants, right handed, native English speakers (eight females; median age, 66 years, MMSE scores $20,22,26\times 3,27,28\times 2,29,30\times 6$)  participated after giving informed written consent. Ethics approval was provided by the Hammersmith Hospital research ethics committee. 

MRI data were obtained on a Philips (Best, The Netherlands) InteraTM $3.0$ Tesla scanner using dual gradients, a phased array head coil, and sensitivity encoding (SENSE) with an undersampling factor of $2$. Functional MR images were obtained using a T2$^{*}$-weighted, gradient-echo, echoplanar imaging (EPI) sequence with whole-brain coverage (repetition time, $2.0 s$; acquisition time, $2.0 s$, echo time, $30 ms$; flip angle, $90^{\circ}$). Thirty-one axial slices with a slice thickness of $3.25 mm$ and an interslice gap of 0.75 mm were acquired in ascending order (resolution, $2.19 \times 2.19 \times 4.0 mm$; field of view, $280 \times 224 \times 128 mm$). Quadratic shim gradients were used to correct for magnetic field inhomogeneities within the anatomy of interest. T1-weighted whole-brain structural images were obtained in all subjects. Stimuli were presented using E-Prime software (Psychology Software Tools, Pittsburgh, PA) run on an IFIS-SA system (In-vivo Corporation, Orlando, FL). 

\subsubsection{fMRI Study Design}
There were two conditions in a blocked design: reading of short sentences (read) and observing Ôfalse fontÕ strings (ff). In the read condition, subjects were required to silently read simple short sentences (containing three words), such as Ôleaves are greenÕ. Each sentence was displayed for $2.25$ seconds followed by another sentence. The acquisition time for each volume was $2$ seconds, so the stimuli were not time-locked to the acquisition onset. The block length was $45$ seconds, allowing $19$ sentences to be displayed per block, as the first $2.25$ seconds of each read block was an instruction screen displaying the words Ôplease read silentlyÕ. During ff trials, subjects observed a string of ff symbols that were arranged in the same form as a three-word sentence, but did not have any meaning, so semantic systems would not be activated during these trials. Again, each ff string was displayed for $2.25$ seconds and therefore, $19$ strings were displayed during one block of ff, and the instruction screen at the start of each block simply displayed false font symbols to inform subjects of the change of task. There were $4$ blocks of read and $4$ blocks of ff, alternating between the two. In total, subjects therefore saw $76$ sentences. Following the scanning, there was a forced-choice recognition memory test outside the scanner, in which subjects had $50\%$ new and $50\%$ old (previously seen in the scanner) sentences. Subjects had to indicate whether they had seen the sentence before (yes or no). The contrast of read with ff was therefore expected to identify systems involved in semantic processing of verbal information and implicit encoding to episodic memory.

\subsubsection{Pre-processing}

The high-resolution T1-structural image was skull stripped using the brain extraction tool (BET) within FSL software~\cite{SmithS:01} to compute the non-linear registration parameters into Montreal Neurological Institute (MNI) standard stereotactic space using FNRIT~\cite{AnderssonJ:08}.

The EPI images were corrected for subject head motion using 6 parameter rigid-body motion correction and performed with MCFLIRT~\cite{JenkinsonM:02}. The corrected data was temporally high pass filtered (Gaussian-weighted LSF straight line subtraction, with $\sigma = 75.0s$~\cite{MarchiniJL:00}, spatially smoothed with a Gaussian Kernel of $3mm$ FHWM, and masked of non-brain voxels using BET. Subsequently, a GLM analysis was performed to estimate the voxel-wise z-scores associated with the linear model that fitted image intensities to stimulus sequence. It is these voxel-wise z-scores together with the subjects' MMSE scores that form, respectively, the output and input variables of the Gaussian Process regression model.

\subsubsection{Gaussian Process Regression}

\begin{figure}[htbp]
\begin{center}
\includegraphics[width=.8\textwidth,height=0.35\textheight]{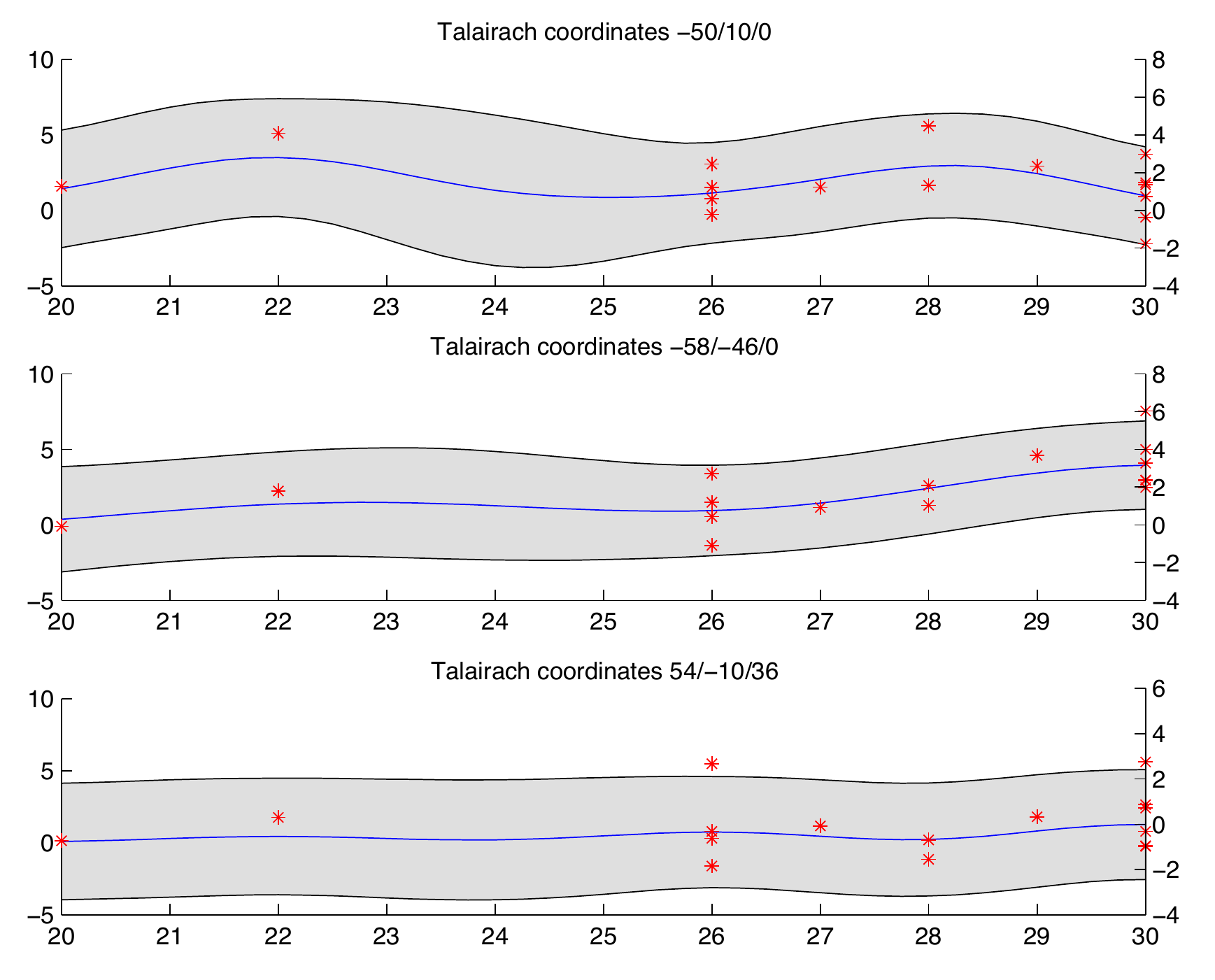}  
\caption{A number of predicted AD activation profiles for different regions in the brain. }
\label{fig:selectedtraces}
\end{center}
\end{figure}

\begin{figure}[htbp]
\begin{center}
\subfloat[The predictive mean]{\label{fig:totmean}\includegraphics[height=0.97\textheight,width=0.48\textwidth]{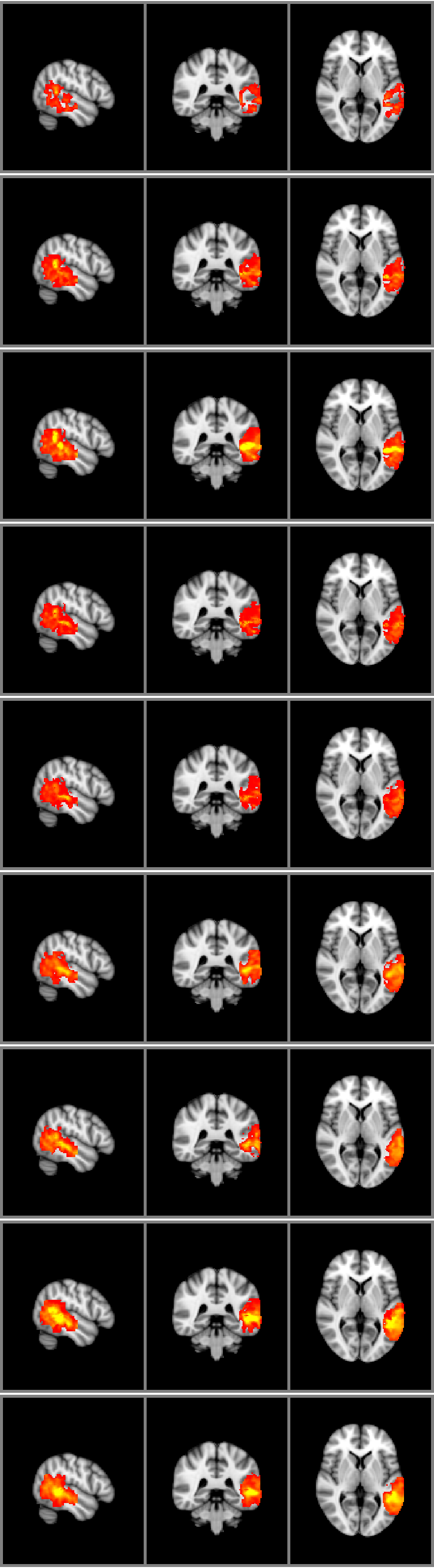}}\quad
\subfloat[The predictive variance]{\label{fig:totvar}\includegraphics[height=0.97\textheight,width=0.48\textwidth]{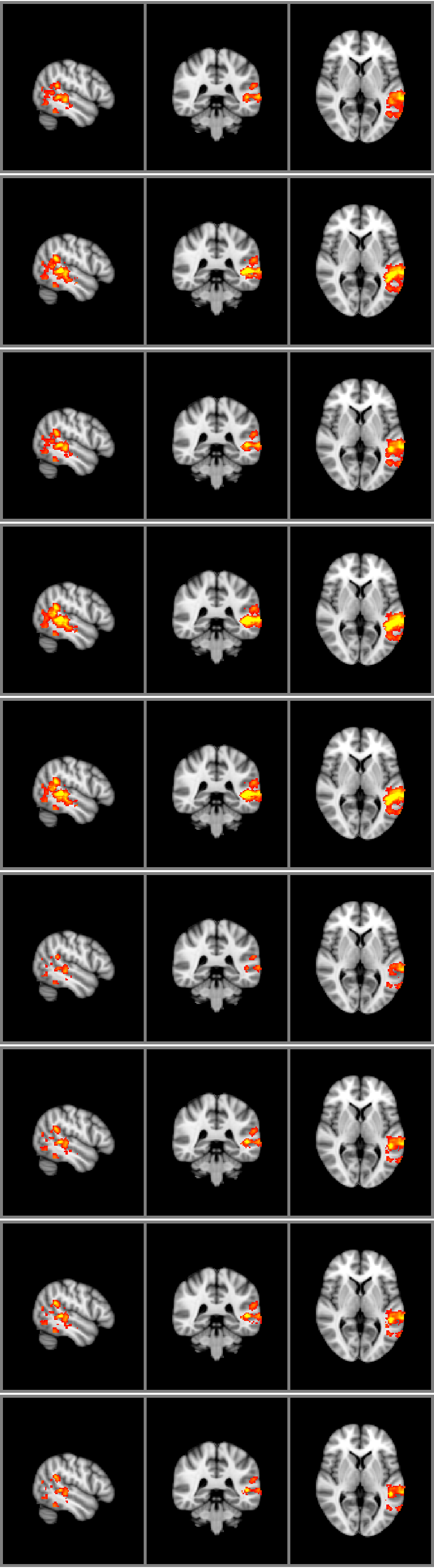}}\quad  
\caption{Changes in temporal lobe activation with improving MMSE. Left: Predicted mean activation; Right: Predicted activation variance.}
\label{fig:totmeanandvar}
\end{center}
\end{figure}

We first qualitatively inspected the predictions made by the GP and its ability to extract a continuous profile of AD image phenotypes. Figure~\ref{fig:selectedtraces} shows a set of example traces collected from various locations of the brain. The asterisks in the figure  show the observed voxel-specific z-scores at their respective MMSE scores. Also shown are the predicted posterior mean activation profile and the predicted posterior variance of the activation profile. The profiles show varying degrees of variability depending on brain region the profile. Some profiles appear to show several peaks along the MMSE axis while others show a delayed and gradual increase at the higher end of the MMSE spectrum. Looking at the spatial distribution of predictions, Figure~\ref{fig:totmeanandvar} shows the changes  predicted activation in the temporal lobe with equidistant increases in MMSE scores, starting with MMSE value of $20$ at top and finishing with MMSE value of $30$ in the bottom row of the figure. Figure~\ref{fig:totmean} shows the predicted mean activation while figure~\ref{fig:totvar} shows the predicted variance in the activation. Clearly visible is the increase in activation spread and intensity. Also very clearly visible is an intermediate increase in variance at MMSE ($4$th and $5$th image row). These images correspond to images at MMSE values that are not contained in the data set. Also, there is a tendency of the variance to decrease with increasing MMSE. This effect is primarily due to the increased quality of the images as patients become more similar to control subjects (MMSE=$30$).

We subsequently investigated quantitatively the performance of the Gaussian Process model with regard to the following questions:
\begin{enumerate}
\item Is non-linear modelling using GPs indeed warranted or would a simpler and linear model suffice?
\item Which brain areas are better modelled using a non-linear description of disease progression?
\item What are the benefits of the GP model compared to traditional multi-level comparisons? 
\end{enumerate}

\begin{figure}[htb]
\begin{center}
\includegraphics[height=0.30\textheight,trim=0 0 35mm 0,clip=false]{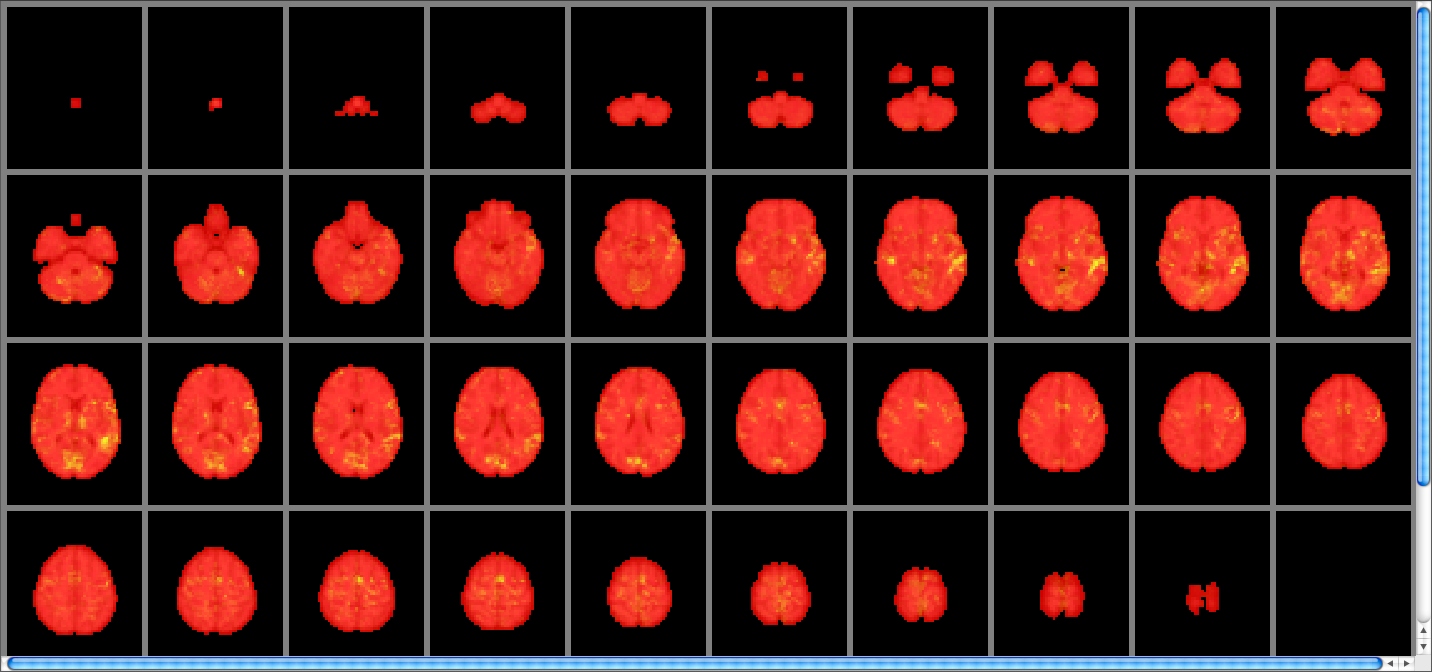}  
\caption{Posterior probability of in favour of a linear increase in activation with MMSE.}
\label{fig:modelprobs}
\end{center}
\end{figure}

To assess which disease profile best described the progression of image phenotypes with MMSE we trained trained two GPs models. The first uses a squared exponential kernel that is capable of non-linear interpolation - see  methods section. The second GP uses a linear kernel (with added noise variance) designed to perform only linear interpolation~\cite{RasmussenCE:06}. Both GPs were trained on identical sets of images. Each GP measures of how well it fitted the data, also known as evidence or marginal log-likelihood, was then extracted. The evidence for the GP with squared exponential covariance function was $-789\times 10^3$ while the evidence of the GP with linear covariance function was $-903\times 10^3$. This suggests a better fit for the  GP with squared exponential covariance function and, hence, for a nonlinear progression of image characteristics with AD's behavioural MMSE score.  Scaled with relative to the number of voxels in the image ($32783$) these evidence values mean that, per voxel, the squared exponential kernel GP is about $3.5$ times more probably than linear kernel GP.

The  difference in evidence is that obtained from that of the entire brain model. However, the evidence of the full brain model can be decomposed into independent and voxel-wise contributions if the Markov random field is fixed at its best estimate. Hence, we can ask the very interesting question as to where the non-linear interpolation model outweighs the linear model?  Figure~\ref{fig:modelprobs} shows the posterior probability in favour of the linear model. Bright/positive values here indicate evidence in favour of the  simpler, linear model while darker/red values indicates a higher probability in favour of the exponential covariance function GP. 

The figure clearly shows an overwhelming tendency towards a nonlinear description of the dependence of image features with MMSE. The largest evidence in favour of the linear model is in the superior and middle temporal gyrus as well as in the occipital regions.  The evidence for the two models in the temporal lobe is  $-243\times 10^3$ and $-505\times 10^3$, respectively for the squared exponential and linear GP. This suggest that for the left temporal lobe in particular the non-linear model is still a better model. However the evidence in favour of the model has dropped to a factor of only $2$ per voxel. Thus, the measured activity in the temporal lobe depends almost monotonically with the MMSE scores. 

Finally, we address the amount of information that is lost by traditional group-based analysis of disease stages compared to a fully interpolating GP model. To be able to use our GP models and mimic such a group comparison, the subjects was allocated into three groups depending on their MMSE scores. Subjects with an MMSE score  below $26$ may be regarded as suffering from AD and allocated a representative MMSE score of $24$ ($5$ Subjects were assigned this group). Subjects with MMSE between $27$ and $29$ were allocated a representative MMSE score of $27$ ($5$ Subjects in this group). Finally all subjects that scored an MMSE of $30$ were part of the  control group in the original design\cite{DhanjalN:10} and retained their MMSE score. The choice of representative scores as guided by the requirement to be equidistant so that GPs were blind to the absolute value of separation between groups. 

\begin{figure}[htb]
\begin{center}
\includegraphics[height=0.35\textheight,trim=0 0 0 0,clip=true]{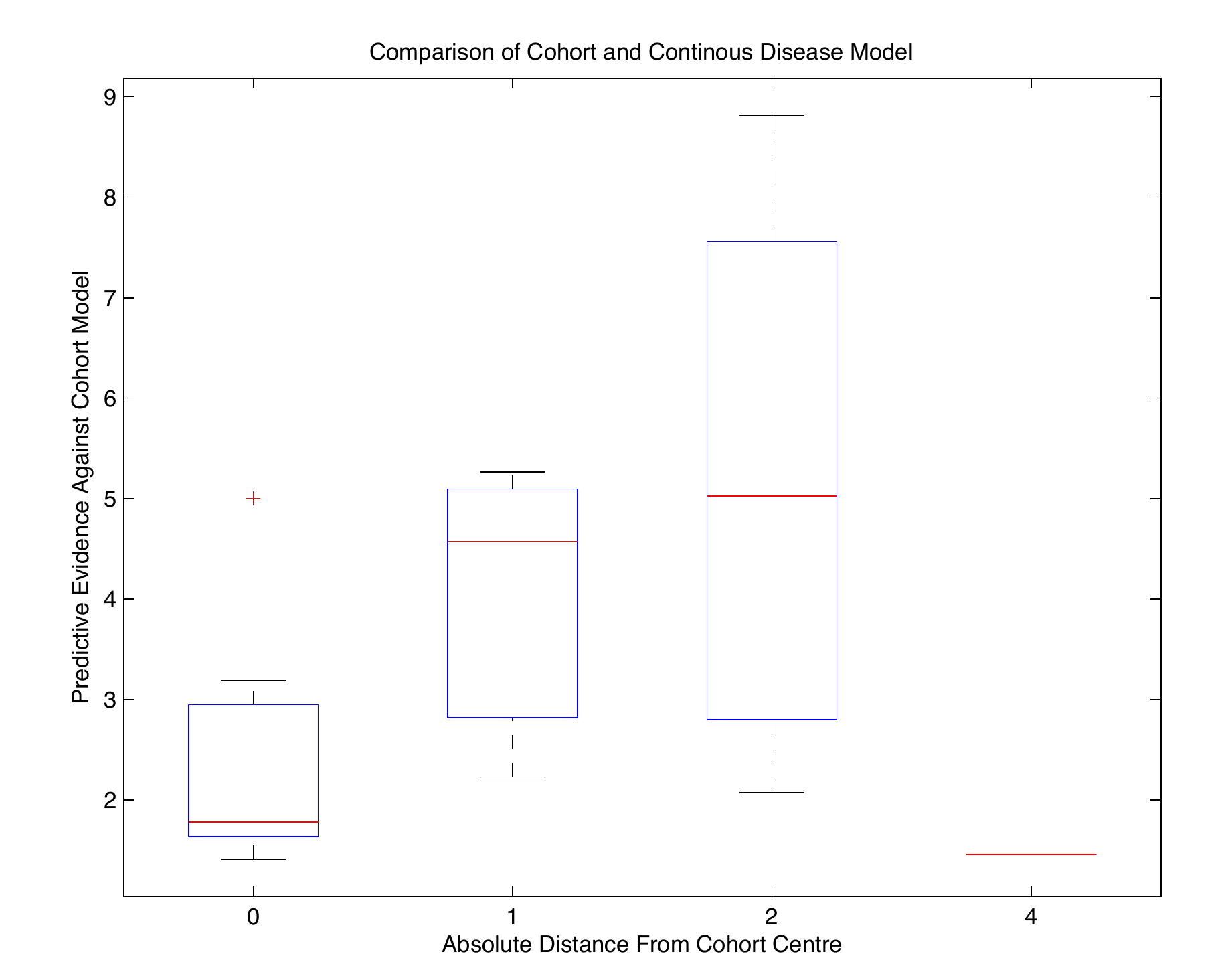}  
\caption{Dependence of model quality on distance to nearest disease stage representative input values. The figure shows the goodness of fit for subject groups with increasing distances from their representative MMSE group score. With only three support points, the discretised model has no evidence to inform the interpolation between the disease stages and so predictions are poor. By contrast, models trained without prior binning of the data have a much greater number of support points. This allows  the model estimate the disease profile and the intermediate stages of the disease much better than a discrete disease stage model.}
\label{fig:cohortcomp}
\end{center}
\end{figure}

We then measured the loss of information due to disease stage grouping by running a leave-one-out cross-validation and to assess the prediction quality. Using the squared exponential GP we have measured an average prediction likelihood of \textcolor{red}{94}. This compares to an  average prediction likelihood of \textcolor{red}{150} for the group-GP. This difference suggests a substantial loss of prediction quality that resulted from the grouping of the data. Further evidence that this  loss is due to binning into disease stages is shown in figure~\ref{fig:cohortcomp}. Clearly visible is the deterioration of the performance of the discrete disease stage model for subjects with greater distance from their representative groups. In contrast, the full regression model describes the intermediate brain function better as it interpolate between the disease states better having been provided with each subjects own behavioural score.  

\section{Discussion}
\label{sec:Discussion}

The main motivation for this research was the observation that numerous fMRI studies into neurological disorders attempt to understand its finer structures using linear regression models in which the explanatory variables are unnecessarily discretised. Often, the sole reason for discretisation is that it allows the use of standard statistical analysis tools, such as ANOVA. 

We have argued that there are a number of key problems associated with this modelling approach. First and foremost, there is no explicit model of the disease process that can describe the changes in neural function in between disease stages. This limitation immediately restricts the amount of detail that can be uncovered. Second, this modelling strategy places low limits in the number of explanatory variables that can be explored due to the need to have sufficient number of measurements in each disease group. Apart from restricting research to the investigation of few disease stages it is also a very inefficient use of the information present in the data. Third, such case-control comparisons approaches lack model-based predictions on the prognosis of the disease. With no explicit model of the disease process no predictions can be made for new cases. Consequently there are no means to verify the disease model or provide vital reliability estimates, e.g. confidence intervals.

Our cross-validation experiments have shown that a continuous model of disease has a better prediction accuracy than standard discrete model. This is to be expected since the grouping of patient population based on MMSE scores destroys any information in the data that could otherwise be used to describe the continuous progression of fMRI features with MMSE. While group-based analysis of the patient population does not make any assumptions about the disease dynamics with respect to behavioural markers, it also unable to make any statements about the disease profile. In contrast, Gaussian process regression does make assumptions about the evolution of fMRI features with MMSE. These assumptions, however,  are very generic assumptions. Thus,  Gaussian process regression is least biased towards a choice of model while also being able to make accurate predictions about unseen cases. 

The Gaussian process regression assumptions may be true or false and, just like any other model. Thus, they must be tested through repeated prediction and validation. In our experiments we have tested two progression hypotheses: a linear increase of z-scores with MMSE and a non-linear dependence of z-scores on MMSE. The linear model  Gaussian process regression resulted in worse model fit scores compared with the squared exponential Gaussian process model - across the whole brain. A more detailed examination showed that the non-linear model does not win outright and that a linear model outperforms the non-linear regression model, for instance, in the temporal lobe of the AD experiment. This suggests that  temporal lobe activity mirrors the performance on the memory tests and that ranking MMSE parallels ranking of  temporal lobe activation intensity. The evidence in favour of the non-linear regression model, however, is wide-spread and overwhelming.

It might be argued that a bootstrap approach should have been used instead of a leave-one out cross-validation. Despite the known bias of this simple form of resampling it is, however, the only choice of resampling technique that is available to our implementation. The reason is our method's computational complexity. A full-brain eave-one-out cross-validation for the 15 subjects in this study runs approximately 2 weeks on a desktop when implemented in Matlab.

A key feature of the model is that approach is reverse of previous use of GPs~\cite{MarquandA:09} and other kernel methods~\cite{SongS:11}. In these studies low-dimensional behavioural data is regressed (or classified) onto a much higher dimensional image feature space. Thus, physical fMRI observations act as explanatory variables, are assumed to be ``controlled'', and used to decode the observed clinical characteristics. Our model, by contrast regresses the high-dimensional  image feature space into behavioural data of much lower dimension, is designed to uncover the neurophysiological differences associated with observed clinical characteristics. In the terminology of~\cite{FristonK:08}, the majority of kernel based methods can be thought of as ``(disease) generative'' models as they model the behavioural characteristics of neuronal activity. By contrast, our model attempts to ``encode'' the brain activity and regards the experimental variable as a cause, not a consequence.

The GP modelling scheme proposed here is naturally applicable whenever the explanatory variables are continuous and the variance (kernel) function is assumed to be smooth. However, the model is not restricted to such conditions and extensions to GPs that account of other, say, categorical input conditions have been proposed~\cite{ShiJQ:07} and can be applied here, also.  Basis function extensions to covariance regression models, for instance, can further reduce the requirements on the smoothness of the variance functions~\cite{HoffP:11}. 

\section{Summary and Possibilities for Improvement}
\label{sec:Summary and Possibilities for Improvement}

To reflect the belief of continuous disease processes we have proposed improvements to disease modelling using Gaussian process regression.  Our approach features a relatively unrestricted and high resolution description of disease processes, e.g. by models describing disease as a continuous spectrum of brain function. By abandoning traditional and information-erosive data binning strategies our model uses all data directly and, thus, makes data utilisation simpler and more efficient.  Finally, a Bayesian description of the disease processes provides full control over model improvement and verification strategies, e.g. through provision of predictive distributions to support the design of clinical studies. 

Our experiments, so far, have been restricted to single input variables. This is far from desirable but was required to achieve robust models in the presence of such small sample sizes (N=15). Also, our focus has been on fMRI features only.   Future work will focus on modelling anatomical measures of AD based on a considerable larger set of behavioural markers and population size. This work will not only serve to confirm the conclusions drawn here, but also bring to the forefront an interesting yet unexplored feature of our model. By determining region-by-region the set of relevant input variables we will be able to ascertain within one model the best set of explanatory variables for each region of the brain. To the best of our knowledge, models with input dependent regional relevance determination have not yet been developed for (f)MRI applications.

\bibliographystyle{apalike}
\bibliography{diseasespectrum}

\end{document}

%% file: userdefs.tex
\usepackage{fullpage}
\usepackage{xspace}
\usepackage{graphicx}
\usepackage{chicago}
\usepackage{color}
\usepackage{rotating}

\makeatletter
\renewcommand{\@seccntformat}[1]{\textsf
  {\csname the#1\endcsname}\hspace{0.5em}}
\renewcommand{\section}{\@startsection
  {section}{1}{0mm}
  {-3.5ex \@plus -1ex \@minus -.2ex}
  {2.3ex \@plus.2ex}
  {\sffamily\Large\bfseries}}
\renewcommand{\subsection}{\@startsection
  {subsection}{2}{0mm}
  {-3.25ex \@plus -1ex \@minus -.2ex}
  {1.5ex \@plus .2ex}
  {\sffamily\large\bfseries}}
\makeatother
\makeatletter
\renewcommand{\@makecaption}[2]{%
  \vspace{5pt}
  {{\small\bf #1:} {\small\sl #2\par}}%
}

\makeatother

\newcommand{\ben}{\begin{enumerate}}
\newcommand{\een}{\end{enumerate}}
\newcommand{\bi}{\begin{itemize}}
\newcommand{\ei}{\end{itemize}}
\newcommand{\bean}{\begin{eqnarray*}}          
\newcommand{\eean}{\end{eqnarray*}}
\newcommand{\bea}{\begin{eqnarray}}
\newcommand{\eea}{\end{eqnarray}}
\newcommand{\ba}{\begin{array}}
\newcommand{\ea}{\end{array}}
\newcommand{\bc}{\begin{center}}
\newcommand{\ec}{\end{center}}
\newcommand{\bt}{\begin{tabular}}
\newcommand{\et}{\end{tabular}}
\newcommand{\bfig}{\begin{figure}[htb]}
\newcommand{\efig}{\end{figure}}

\newcommand{\thrdm}[1]{3{\tiny $^{ \mbox{rd}}$}}
\newcommand{\fothm}[1]{4{\tiny $_{ \mbox{th}}$}}
\newcommand{\eqnref}[1]{{~(\ref{#1})}}

\newcommand{\inv}{{^{\mathsf{-1}}}}

\newcommand{\tp}{^{\scriptscriptstyle \mathsf T}}

\newcommand{\argmax}[1]{{\underset{#1}{\mbox{argmax}}}}

%% file: diseasespectrum.bbl
\begin{thebibliography}{}

\bibitem[Andersson et~al., 2008]{AnderssonJ:08}
Andersson, J., Smith, S., and Jenkinson, M. (2008).
\newblock Fnirt - fmrib's non-linear image registration tool.
\newblock In {\em Fourteenth Annual Meeting of the Organization for Human Brain
  Mapping}.

\bibitem[Besag, 1974]{BesagJ:74}
Besag, J. (1974).
\newblock Spatial interaction and the statistical analysis of lattice systems.
\newblock {\em Journal of the Royal Statistical Society, Series B},
  32(2):192--236.

\bibitem[Besag, 1986]{BesagJ:86}
Besag, J. (1986).
\newblock On the statistical analysis of dirty pictures.
\newblock {\em Journal of the Royal Statistical Society Series B},
  48(3):259--302.

\bibitem[Dhanjal et~al., 2010]{DhanjalN:10}
Dhanjal, N., Warren, J., and Wise, R. (2010).
\newblock Verbal memory decline in alzheimer's disease is associated with an
  impairment of frontal executive control.
\newblock In {\em Proceedings of the American Academy of Neurology Annual
  Meeting}.

\bibitem[Friston et~al., 2008]{FristonK:08}
Friston, K., Chu, C., and Mour{\~a}o-Miranda, J. (2008).
\newblock Bayesian decoding of brain images.
\newblock {\em NeuroImage}, 39:181--205.

\bibitem[Hoff and Niu, 2011]{HoffP:11}
Hoff, P. and Niu, X. (2011).
\newblock A covariance regression model.
\newblock {\em to appear in Statistica Sinica}.

\bibitem[Huaien and Puthusserypady, 2004]{HuaienL:04}
Huaien, L. and Puthusserypady, S. (2004).
\newblock Bayesian radial basis function network for modeling fmri data.
\newblock In {\em Proceedings of the IEEE Engineering in Medicine and Biology
  Society Conference}, pages 450--453.

\bibitem[Jenkinson et~al., 2002]{JenkinsonM:02}
Jenkinson, M., Bannister, P., and Smith, S. (2002).
\newblock Improved optimisation for the robust and accurate linear registration
  and motion correction of brain images.
\newblock {\em NeuroImage}, 17(2):825--841.

\bibitem[Lange et~al., 1999]{LangeN:99}
Lange, N., Strother, S., Anderson, J., Nielsen, F., Holmes, A., Kolenda, T.,
  Savoy, R., and Hansen, L. (1999).
\newblock Plurality and resemblance in fmri data analysis.
\newblock {\em NeuroImage}, 10(3):282--303.

\bibitem[MacKay, 2003]{MacKay:2003}
MacKay, D. (2003).
\newblock Information theory, inference and learning algorithms.
\newblock {\em Cambridge University Press}.

\bibitem[Marchini and Ripley, 2000]{MarchiniJL:00}
Marchini, J. and Ripley, B. (2000).
\newblock A new statistical approach to detecting significant activation in
  functional mri.
\newblock {\em NeuroImage}, 12(4):366--380.

\bibitem[Marquand et~al., 2010]{MarquandA:10}
Marquand, A., Howard, M., Brammer, M., C, C.~C., Coen, S., and
  Mour{\~a}o-Miranda, J. (2010).
\newblock Quantitative prediction of subjective pain intensity from whole-brain
  fmri data using gaussian processes.
\newblock {\em Neuroimage}, 49(3):2178--89.

\bibitem[Marquand et~al., 2009]{MarquandA:09}
Marquand, A.~F., Howard, M., Brammer, M.~J., and Mourao-Miranda, J. (2009).
\newblock Probabilistic classification of functional magnetic resonance imaging
  (fmri) data using gaussianprocess classification: Application to pain
  perception.
\newblock In {\em Proceedings of the 15th Annual Meeting of the Organisation
  for Human Brain Mapping}.

\bibitem[Martino et~al., 2010]{DeMartinoF:10}
Martino, F.~D., de~Borst, A., G, G.~V., Goebel, R., and Formisano, E. (2010).
\newblock Predicting eeg single trial responses with simultaneous fmri and
  relevance vector machine regression.
\newblock {\em NeuroImage}.

\bibitem[McKhann et~al., 1984]{McKhannG:84}
McKhann, G., Drachman, D., Folstein, M., Katzman, R., Price, D., and Stadlan,
  E. (1984).
\newblock Clinical diagnosis of alzheimer's dis- ease: report of the
  nincds-adrda work group under the auspices of the department of health and
  human services task force on alzheimer's disease.
\newblock {\em Neurology}, 34:939--944.

\bibitem[Neal, 1997]{NealR:97}
Neal, R. (1997).
\newblock Monte carlo implementation of gaussian process models for bayesian
  regression and classification.
\newblock Technical Report 9702, University of Toronto.

\bibitem[Nestor et~al., 2004]{NestorP:04}
Nestor, P., Scheltens, P., and JR, J. (2004).
\newblock Advances in the early detection of alzheimer's disease.
\newblock {\em Nat Med}, 10:S34--41.

\bibitem[Ni et~al., 2008]{NiY:08}
Ni, Y., Chu, C., Saunders, C., and Ashburner, J. (2008).
\newblock Kernel methods for fmri pattern prediction.
\newblock In {\em International Joint Conference on Neural Networks}.

\bibitem[Rasmussen and Quinonero-Candela, 2005]{RasmussenCE:05}
Rasmussen, C. and Quinonero-Candela, J. (2005).
\newblock Healing the relevance vector machine through augmentation.
\newblock In {\em Proceedings of the 22nd international conference on Machine
  learning}, pages 689--696.

\bibitem[Rasmussen and Williams, 2006]{RasmussenCE:06}
Rasmussen, C. and Williams, C. (2006).
\newblock {\em Gaussian Processes for Machine Learning}.
\newblock MIT Press.

\bibitem[Rezek et~al., 2010]{RezekI:10}
Rezek, I., Dhanjal, N., and Wise, R. (2010).
\newblock Models of disease spectra.
\newblock In {\em Proceedings of the 16th Annual Meeting of the Organization
  for Human Brain Mapping}.

\bibitem[Shi et~al., 2007]{ShiJQ:07}
Shi, J.~Q., Wang, B., Murray-Smith, R., and Titterington, D.~M. (2007).
\newblock Gaussian process functional regression modeling for batch data.
\newblock {\em Biometrics}, 63(3):714--723.

\bibitem[Smith et~al., 2001]{SmithS:01}
Smith, S., Bannister, P., Beckmann, C., Brady, M., Clare, S., Flitney, D.,
  Hansen, P., Jenkinson, M., Leibovici, D., Ripley, B., Woolrich, M., and
  Zhang, Y. (2001).
\newblock Fsl: New tools for functional and structural brain image analysis.
\newblock In {\em Seventh Int. Conf. on Functional Mapping of the Human Brain}.

\bibitem[Song et~al., 2011]{SongS:11}
Song, S., Zhan, Z., Long, Z., Zhang, J., and Yao, L. (2011).
\newblock Comparative study of svm methods combined with voxel selection for
  object category classification on fmri data.
\newblock {\em PLoS ONE}, 6(2):e17191.

\bibitem[Stein, 1999]{SteinML:99}
Stein, M. (1999).
\newblock {\em Interpolation of Spatial Data: Some Theory for Krieging}.
\newblock Springer Verlag.

\bibitem[Yetkin et~al., 2006]{YetkinFZ:06}
Yetkin, F.~Z., Rosenberg, R., Weiner, M., Purdy, P., and Cullum, C.~M. (2006).
\newblock Fmri of working memory in patients with mild cognitive impairment and
  probable alzheimer's disease.
\newblock {\em European Radiology}, 16(1):193--206.

\end{thebibliography}
